\documentclass{article}

\usepackage{arxiv}

\usepackage[utf8]{inputenc} 
\usepackage[T1]{fontenc}    
\usepackage{hyperref}       
\usepackage{url}            
\usepackage{booktabs}       
\usepackage{amsfonts}       
\usepackage{nicefrac}       
\usepackage{graphicx}
\usepackage{natbib}
\usepackage{doi}
\usepackage{array}
\newcolumntype{C}[1]{>{\centering}m{#1}}
\newcolumntype{M}[1]{>{\centering\arraybackslash}m{#1}}
\usepackage{multirow}
\usepackage{float}
\usepackage{caption}
\usepackage{subcaption}
\usepackage{amssymb}
\usepackage{amsmath}
\usepackage{setspace}
\usepackage{algorithm}
\usepackage{algpseudocode}
\usepackage{xcolor}

\makeatletter
    \def\tagform@#1{\maketag@@@{\normalsize(#1)\@@italiccorr}}
\makeatother

\title{Rethinking Selection in Generational Genetic Algorithms to Solve Combinatorial Optimization Problems: An Upper Bound-based Parent Selection Strategy for Recombination}

\author{Prashant Sankaran
\\
	Department of Industrial and Systems Engineering\\
	University at Buffalo\\
	Buffalo, NY 14260 \\
	\texttt{psankara@buffalo.edu} \\
	\And
	Katie McConky \\
	Department of Industrial and Systems Engineering\\
	Rochester Institute of Technology\\
	Rochester, NY \\
	\texttt{katie.mcconky@rit.edu} 
}

\date{}

\renewcommand{\shorttitle}{Upper Bound Based-Parent Selection}

\hypersetup{
pdftitle={Rethinking Selection in Generational Genetic Algorithms to Solve Combinatorial Optimization Problems: An Upper Bound-based Parent Selection Strategy for Recombination},
pdfsubject={CoRR.NEC, CoRR.AI},
pdfauthor={Prashant Sankaran, Katie McConky},
pdfkeywords={First keyword, Second keyword, More},
}

\begin{document}
\maketitle

\begin{abstract}
Existing stochastic selection strategies for parent selection in generational GA help build genetic diversity and sustain exploration; however, it ignores the possibility of exploiting knowledge gained by the process to make informed decisions for parent selection, which can often lead to an inefficient search for large, challenging optimization problems. This work proposes a deterministic parent selection strategy for recombination in a generational GA setting called Upper Bound-based Parent Selection (UBS) to solve NP-hard combinatorial optimization problems. Specifically, as part of the UBS strategy, we formulate the parent selection problem using the MAB framework and a modified UCB1 algorithm to manage exploration and exploitation. Further, we provided a unique similarity-based approach for transferring knowledge of the search progress between generations to accelerate the search. To demonstrate the effectiveness of the proposed UBS strategy in comparison to traditional stochastic selection strategies, we conduct experimental studies on two NP-hard combinatorial optimization problems: team orienteering and quadratic assignment. Specifically, we first perform a characterization study to determine the potential of UBS and the best configuration for all the selection strategies involved. Next, we run experiments using these best configurations as part of the comparison study. The results from the characterization studies reveal that UBS, in most cases, favors larger variations among the population between generations. Next, the comparison studies reveal that UBS can effectively search for high-quality solutions faster than traditional stochastic selection strategies on challenging NP-hard combinatorial optimization problems under given experimental conditions.
\end{abstract}


\section{Introduction}
Genetic algorithms (GAs) are stochastic population-based metaheuristics inspired by natural evolution \citep{Talbi2009}. Over the years, GAs have been widely adopted as approximate solution approaches to solve challenging real-world problems \citep{Ghaheri2015, Yang2014} owing to the conceptual simplicity, versatility, and parallelizability \citep{Sivanandam2008, Yang2014}.

Typically, in GAs, evolutionary operations such as selection, recombination, and mutation are iteratively applied to existing individuals in a population to create new, potentially better individuals \citep{Sivanandam2008}, while a replacement is periodically applied to systematically replace existing individuals with new individuals in a population \citep{Talbi2009}. The intuition behind the recombination, mutation, selection, and replacement operators are as follows. The recombination operator is used to fuse individuals within a population to form new individuals, mutation to add diversity to the population, selection to direct the search toward promising solution regions \citep{Back1994, Blickle1996}, and replacement to maintain a manageable population size \citep{Sivanandam2008}.

For search algorithms like GA, there needs to be a balance between exploration (or diversification) and exploitation (or intensification) to ensure adequate search progress \citep{Crepinsek2013}. In GAs, the recombination, mutation, and selection operators together control the extent of the tradeoff between the exploration and exploitation of the search space \citep{Back1994}. Specifically, the recombination and mutation operators are inclined towards exploration, while selection is towards exploitation. However, selection, capable of steering the search direction, is more responsible for balancing exploration and exploitation \citep{Back1994}. Therefore, it is critical to have a good selection strategy that can effectively balance exploration and exploitation.

Typically, selection operator design focuses on controlling the selection pressure to balance between exploration and exploitation. Selection pressure determines the extent of bias in choosing individuals with a better fitness value for recombination \citep{Back1994, Rogers1999}. Historically, diversity-preserving strategies and stochasticity have been used to adjust selection pressure \citep{Reeves2010, Rogers1999, Srinivas1994}. Although relying on stochasticity helps build genetic diversity and exploration, it ignores the possibility of exploiting knowledge gained by the process to make informed decisions for parent selection, thereby leading to an inefficient search.

In this article, we propose using a deterministic parent selection strategy called upper bound-based parent selection (UBS) that considers the entire population and knowledge gained about the search progress from the previous generation while selecting parents for recombination within the generational GA framework. Further, we hypothesize that the combined effect of upper bound-based greedy parent selection and sharing search progress between generations within the UBS strategy can help find high-quality solutions faster than traditional stochastic selection strategies on challenging NP-hard combinatorial optimization problems like the team orienteering problem \citep{Chao1996} and the quadratic assignment problem \citep{Koopmans1957}.

The underlying UBS strategy transforms the parent selection problems into a multi-armed bandit (MAB) problem and solves using a modified variant of the deterministic upper confidence bound-based reinforcement learning approach \citep{Auer2002,AuerFischer2002}. The rationale behind using a reinforcement learning approach to solve parent selection in GA lies in the similarity of the exploration and exploitation tradeoff dilemma faced by the MAB problem \citep{Slivkins2019, Sutton2018} and GA's selection problem. Further, by incorporating reinforcement learning into the selection process, we aim to develop an intelligent and efficient search algorithm capable of addressing large-scale problems with limited computational resources and time. {The main contributions of this work are as follows.
\begin{enumerate}
    \item A deterministic learning-based parent selection strategy that addresses the exploration and exploitation tradeoff dilemma.
    \item The parent selection problem within the generational GA framework is solved using the MAB framework.
    \item Introduction of knowledge transfer about the search progress between generations using a new similarity-based approach.
    \item UBS and several traditional selection strategies are characterized and compared on NP-hard combinatorial optimization problems such as team orienteering and quadratic assignment problems within the generational GA framework.
\end{enumerate}

The rest of the article is structured as follows: Section 2 provides a brief background on simple GA and the MAB problem, and Section 3 provides some related work that covers the selection strategies in GAs with a particular interest in the selection for recombination and the MAB framework, Section 4 presents our proposed selection strategy, Section 5 provides our setup for the characterization and the comparison experiments, and Section 6 covers the results on the team orienteering \citep{Chao1996} and quadratic assignment \citep{Burkard1997} benchmark problem dataset from the literature. Section 7 shares our conclusions, and Section 8 closes with the limitations and future work.

\section{Background}\label{background}
In this section, we will briefly cover the necessary background on GAs and MABs and their upper confidence bound-based solution approaches that are relevant to our work.

\subsection{Genetic Algorithms}

GAs are metaheuristics for searching a large solution space by evolving a starting population of feasible solutions over several generations to solve optimization problems approximately \citep{Whitley2019}. The key evolutionary process involves selection, recombination, and mutation \citep{Reeves2010,Whitley2019}. Further, additional strategies, such as elitism, can be augmented to ensure the best set of solutions appears in the next generation of the population \citep{Whitley2019}. In GAs, solutions or population individuals are called chromosomes, and the solution's elements (such as letters of a sequence) are called genes \citep{Sivanandam2008}. Further, we can differentiate chromosomes within a given generation as parent or offspring depending on the state of the chromosome. In a given generation, chromosomes in the population that undergo variation operations like recombination and mutation are called the parent chromosomes. The resultant chromosomes obtained from applying variations to parent chromosomes are offspring chromosomes. Next, we briefly go over how GAs work.

Since numerous variations of GAs are available in the literature, we describe a version that applies throughout this work. The overall generational GA workflow can be understood as follows. First, randomly initialize a population of chromosomes and evaluate each chromosome using a fitness function. Typically, we use the objective value of the optimization problem as the fitness function \citep{Azarbonyad2014,Ferreira2014}. We now move on to the selection operation.

In selection, we select a subset of the population to form an intermediate population based on a predefined strategy, for example, proportional selection \citep{Blickle1996}, where the relative frequency of selected chromosomes is proportional to the fitness value. Next, once we have our intermediate population, we enter the parent selection operation, where we choose a subset of parent chromosomes based on an appropriate strategy, usually uniform random, and apply the recombination operation with a probability of $p_r$, and subsequently, mutation with a probability of $p_m$ \citep{Whitley2019}. Typically, in recombination, two or more chromosome sections, or partial gene sequences, are selected randomly and exchanged among the different parent chromosomes to form unique offspring that share characteristics of each parent \citep{Whitley2019}. Further, these chromosomes undergo mutation, which involves random gene swaps, deletion (if applicable), and addition \citep{Azarbonyad2014, Ferreira2014}. At the end of recombination and mutation, we obtain offspring chromosome(s) for each parent chromosome pair. These offspring then undergo evaluation, and only the best-fit chromosomes, determined by predefined criteria, become part of the new population for the next generation, along with a predetermined subset of the old population having the best fitness scores, known as the elite subgroup \citep{Ferreira2014}. Until we meet the desired termination condition, such as maximum generations, elapsed time, or no change in fitness \citep{Sivanandam2008}, the cycle repeats, and the best solution found so far is the algorithm's output. We refer readers to \citep{Kramer2017, Sivanandam2008} for more comprehensive coverage of GAs.

\subsection{Multi-Armed Bandits}
In an MAB problem, given a set of arms (or actions) with an unknown distribution of rewards in a single-state Markov decision process, the goal is to maximize the cumulative rewards of selecting arms over a given time horizon $T$ \citep{Sutton2018}. Further, effectively solving the MAB problem involves addressing the tradeoff between exploring new arms to find better rewards and exploiting the current best arms to reap immediate benefits. If the unknown distribution of rewards is stationary, a simple average-based action-value estimates approach is sufficient \citep{Sutton2018}. However, sophisticated approaches are necessary since most real-world problems are nonstationary, meaning that each arm's reward distribution may change over time. To this end, the upper confidence bound-based approach proposed in \citep{Auer2002, AuerFischer2002} is a good solution. The key idea behind the upper confidence bound algorithm (UCB1) for solving the MAB problem is to take an optimistic view in the face of uncertainty in action-value estimates of the arms. Precisely, in the UCB1 algorithm, we select arms based on~\eqref{eq2.1}

\begin{equation}
    \label{eq2.1}
    A_t = \arg_a \max [Q_t(a) + c\sqrt{\frac{\log(T)}{N_t(a)}}],
\end{equation}

\noindent where $A_t$ denotes the selected arm at time step $t$, $Q_t (a)$ denotes the action-values estimate for arm $a$, and the square root term denotes the upper bound on the action-values estimates. Further, $c$ controls the scale of the upper bound. Lastly, $N_t(a)$ denotes the number of times action a is selected until time step $t$.

It is evident from~\eqref{eq2.1} that as we select an arm more frequently, the upper bound value, square root term, decays to presumably indicate that the action values estimates are approaching their true values. Although UCB1 can solve MAB problems, it can get quite challenging when the number of arms grows substantially. A particular case of interest to our work includes the continuum-armed bandits that satisfy the Lipschitz condition \citep{Slivkins2019}. In the continuum-armed bandits or the general problem often referred to as the Lipschitz bandit problem, we consider the arms to lie in a known metric space ($X, D$), where $X=[0,1]$ is a ground set, and $D(x,y)=L\cdot |x-y|$ is the metric, and $L$ is a Lipschitz constant. To address this problem, an extension of the UCB1 algorithm called the zooming algorithm is proposed by \citep{Kleinberg2008}. The zooming algorithm combines the upper confidence bounds approach from UCB1 with an adaptive discretization approach. The key idea in the adaptive zooming algorithm is to place probes or arms through an iterative process to eventually identify regions of higher payoffs. To this end, the algorithm begins by activating an arm not already covered by a confidence radius of other active arms. Then, the algorithm plays an active arm using the section rule in~\eqref{eq2.2}. Next, the reward estimates ($\bar{\mu_t}$) and confidence radius ($r_t$) are updated accordingly. As we update the radius, new arms previously covered by the radius may get uncovered for future selection, providing the zoom-in effect. Finally, the algorithm continues for a fixed time horizon $T$.

\begin{equation}
    \label{eq2.2}
    Index_t (x) = \bar{\mu_t}(x) + 2r_t(x).
\end{equation}

In~\eqref{eq2.2}, $Index_t (x)$ denotes the index of the arm with the highest upper bound, $\bar{\mu_t}$ is the average rewards until time $t$, and $r_t (x)$ denotes the confident radius defined in~\eqref{eq2.3}, where $n_t (x)$ denotes the number of times action $x$ is selected until time step $t$.

\begin{equation}
    \label{eq2.3}
    r_t(x) = \sqrt{\frac{2 \log(T)}{n(x)+1}}.
\end{equation}

For more comprehensive coverage of the multi-arm bandit problems and the solution approaches presented here, we refer readers to \citep{Auer2002, AuerFischer2002, Slivkins2019} for the upper confidence bound algorithm and \citep{Kleinberg2008, Kleinberg2019} for the zooming algorithm.

\section{Related Work}\label{relatedwork}
The specific aspect addressed in this work on parent selection for recombination in generational GAs using an MAB framework is unexplored. Therefore, we expand our search for work addressing parent selection for recombination without an intermediate population or parent selection using an MAB framework or evolutionary algorithm using the MAB framework.

The GA community has explored parent selection strategies for recombination in the past. \citep{Whitley1988} proposed the GENITOR algorithm, which employed a one-parent-selection-at-a-time approach and modeled the first steady-state GA \citep{Whitley2019}. In steady-state GA, the selected parents directly undergo recombination and other evolutionary operations instead of selecting the entire population for each generation. Later, after performing all the evolutionary operations, the offspring are added back into the population. Further, the chromosome with the worst fit is eliminated, thereby maintaining the population size across evolutions. The process continues until a predefined termination condition. Stead-state GA is known to improve the population monotonically; therefore, it has higher selection pressure and quicker loss of diversity \citep{Whitley2019}.

Further, \citep{Thierens1994} proposed a GA with elitist recombination, which combines selection and recombination into a single-step evolution process. The algorithm works as follows. For each generation, the algorithm first shuffles the population. Then, for each parent pair allowed to mate, two offspring are generated to form a family. Next, the algorithm conducts a tournament selection for each family and selects the top two chromosomes based on the fitness function while discarding the remaining chromosomes. The authors report that the elitist recombination resembles a standard tournament selection algorithm with a tournament size of two. Additionally, they conclude that their approach is insensitive to population size compared to tournament selection and the choice of the recombination probability.

\citep{Matsui1999} explored the benefits of similarity-based selection for recombination. Specifically, they introduce correlative tournament selection, a variant of ordinary tournament selection, where the tournament selects the first parent, and among a randomly sampled subpopulation, one having the highest correlation with the first parent is the second parent. Furthermore, \citep{Matsui1999} proposes correlative family-based selection, a strategy to determine if a given chromosome makes it to the next generation, based on \citep{Thierens1994}'s elitist recombination. In correlative family-based selection, the best-fit chromosome from a family of 4 and the least correlated to the best-fit chromosome are selected. They evaluate the correlative family-based selection approach on the Royal Road and nonstationary knapsack problems. The experimental results indicate that the proposed approach improves population diversity and is better than the traditional tournament selection strategy.

Next, in the context of selection using the MAB framework for parent selection, we found \citep{Loshchilov2011} to best represent this research avenue. \citep{Loshchilov2011} proposed a MAB-based steady-state parent selection approach for solving multi-objective problems using an evolutionary framework different from the GA framework considered in this work. The proposed evolutionary framework is a variant of the covariance matrix adaptation evolution strategy (CMA-ES). Further, they evaluate multiple reward functions considering both the parent and offspring. Then, they assign the same rewards to all chromosomes involved. The underlying MAB solution used in \citep{Loshchilov2011} consists of a reward function to encourage exploitation and a time-windowed approach to exploration. Precisely, each parent gets to be selected at least once within $x$ time steps since it was chosen last. Here, $x$ denotes the length of the time window. Further, they evaluate their approach in two well-known bi-objective problems from the literature. Finally, they report that their proposed evolutionary selection strategies using an MAB framework yield practical speed-up on the evaluated bi-objective unimodal problems while performing subpar on bi-objective multi-modal problems.

Lastly, several researchers have used the MAB framework to address the adaptive operator selection problem in evolutionary algorithms \citep{Belluz2015, DaCosta2008, Goncalves2015, Li2014}. The pioneering work on adaptive operator selection using the MAB framework is due to \citep{DaCosta2008}. \citep{DaCosta2008} propose a new dynamic MAB-based approach (D-MAB) to solve the adaptive operator selection problem in evolutionary algorithms. Specifically, the D-MAB approach hybridizes the UCB1 algorithm with a statistical test called the Page-Hinkley test to detect significant changes in the population or environment and accordingly restart the UCB1 algorithm. Further, within D-MAB's UCB1 algorithm, numerical rewards are assigned based on the performance of selected operators in producing high-quality individuals. They evaluate D-MAB on three artificial scenarios: the Boolean, Uniform, and Outlier. The results indicate competitive performance against probability matching and adaptive pursuit methods, while a simple UCB1-based approach was also sufficient on the Boolean. Further, \citep{Belluz2015} proposed an improved D-MAB approach to address the need for parallel evaluation and operator management. The results on the synthetically generated One Max problems indicate better performance in cases involving a larger variety of operators. Next, in \citep{Li2014}, a fitness-rate-ranked-based MAB algorithm is proposed to address the adaptive operator selection problem within the multi-objective evolutionary algorithm based on decomposition (MOEA/D). The experimental results on several multi-objective benchmark problems indicate robust operator selection performance. Lastly, building on \citep{Li2014}, \citep{Goncalves2015} propose two algorithms based on UCB1: MOEA/D-UCB-Tuned and MOEA/D-UCB-V, which uses the variance of the operators' rewards to balance the exploration and exploitation tradeoff. The experimental results indicate significant performance improvement over several approaches, including the method proposed in \citep{Li2014}.

\subsection{Discussion}
Although \citep{Loshchilov2011} use an MAB framework like our proposed approach, they only select one arm at a time while we select two. Some key limitations in selecting one arm at a time instead of two arms considering a generational GA case include the following. Firstly, single-arm selection increases the risk of selecting the same parent again in cases when the upper bounds have decayed. Secondly, when selecting only one arm at a time, the exploration can be slower, thereby computationally less efficient on large problem instances. Third, single-arm selection can reduce selection pressure, thereby delaying convergence.

Furthermore, they assign the same rewards to both the parents and the offspring. In contrast, we use the fitness score directly for parent and offspring rewards. Note that we only compute the offspring rewards after generating all the offspring.

Also, \citep{Loshchilov2011} use a time-windowed approach to encourage exploration instead of an upper bound approach and a similarity-based approach to initialize the upper bounds for the offspring used in our proposed UBS approach. Although the time-windowed approach ensures fair exploration, it creates undue pressure on the exploration of parents as their time window nears closure. Further, such a time-windowed approach cannot be implemented in a generational GA setting since the population might differ, moving from generation to generation.

Lastly, they evaluate their parent selection approach for generating offspring using a steady-state evolutionary framework on multi-objective problems. However, we evaluate our approach on a typical GA workflow using a single objective combinatorial optimization problem.

\section{Upper Bound-Based Parent Selection Strategy}\label{UBS}

This section presents our proposed selection strategy for genetic algorithms called UBS. The essence of the UBS strategy is to combine the selection pressure induced by the fitness function from GA for exploitation along with the upper bound concept from MAB literature to encourage a deterministic, structured exploration of the search space. To this end, in the UBS strategy, we formulate and solve selection for recombination in GAs as an MAB problem \citep{Slivkins2019, Sutton2018}. 

In a typical MAB problem, given a set of arms (or actions) with an unknown distribution of rewards in a single-state Markov decision process, the goal is to maximize the cumulative rewards of selecting arms over a given time horizon \citep{Sutton2018}. Further, effectively solving the MAB problem involves addressing the tradeoff between exploring new arms to find better rewards and exploiting the current best arms to reap immediate benefits, similar to the dilemma faced by the selection process in GAs. Therefore, formulating selection in GAs as an MAB problem extends naturally with a few adjustments.

To formulate selection in GAs as an MAB problem, we first map the chromosomes in GAs to the arms of a bandit. Next, instead of selecting one arm at each selection step, as usually done in MAB, we choose two arms corresponding to two parent chromosomes. To solve this MAB-based selection problem, we propose a modified upper confidence bound algorithm \citep{Auer2002, AuerFischer2002}. However, a few key challenges exist. First, the same chromosomes may or may not appear in future generations. Secondly, in MAB problems, we explore each arm at least once before determining the expected rewards for future selections. However, in GAs, this need not be the case. Next, in MAB problems, the rewards are typically a function of selected arms; however, in GAs, selection operation first creates an intermediate population, which serves as a pool for selecting the chromosomes for applying recombination and mutation operations. Lastly, in GAs, typically, we use the fitness function to indicate the quality of a chromosome rather than rewards.

To overcome the first two challenges, we propose two rules for initializing/updating the upper confidence bound on rewards: one for initializing new arms introduced in each generation and one for each successive top-2 selection within a generation. Here, the number of arms in each generation and the number of selection (or recombination) steps within each generation are all assumed to be predefined parameters. Now, solving the second challenge with an online update rule provides the opportunity to eliminate the need for an intermediate population separating selection and recombination operations in traditional GA workflow, thereby resolving the third challenge. Lastly, we propose to use a ranked-based fitness function that preserves diversity based on \citep{Vidal2022} as parent rewards. Some of the key reasons behind using a fitness function as a reward function include enabling ease of computation and capitalizing on the knowledge of the fitness function from the GA community. Figure~\ref{fig4.1}  shows the new GA workflow. In the following few subsections, we dive deeper into the actual working of UBS in this GA workflow.

\begin{figure}[h]
    \begin{center}
        \includegraphics[width=0.4\linewidth]{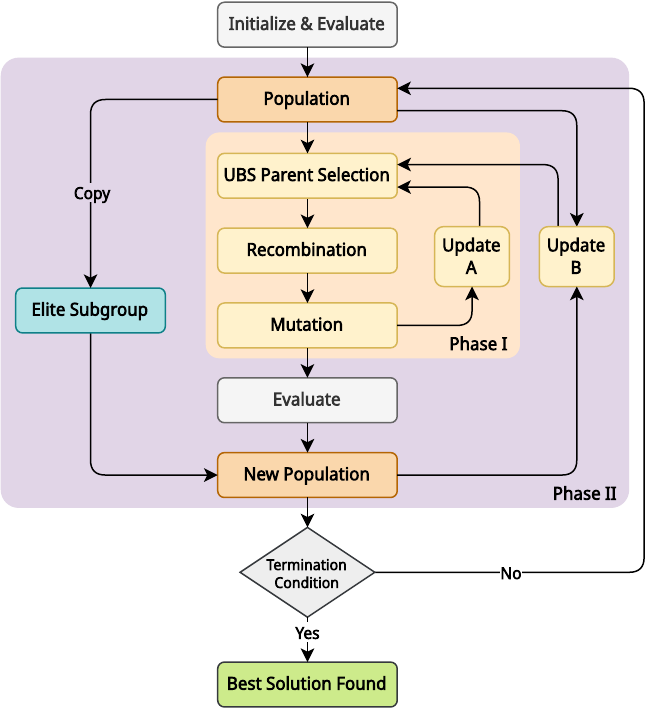}
    \end{center}
    \caption{{Modified workflow for a genetic algorithm with UBS.}}
    \label{fig4.1}
\end{figure}

\subsection{GA with Upper Bound-based Selection (UBS)}
As presented in Figure~\ref{fig4.2}, the overall UBS strategy contains two phases: within a generation (Phase I) and at the start of a new generation (Phase II). We describe each phase in subsequent sections.

\begin{figure}[h]
    \begin{center}
        \includegraphics[width=\linewidth]{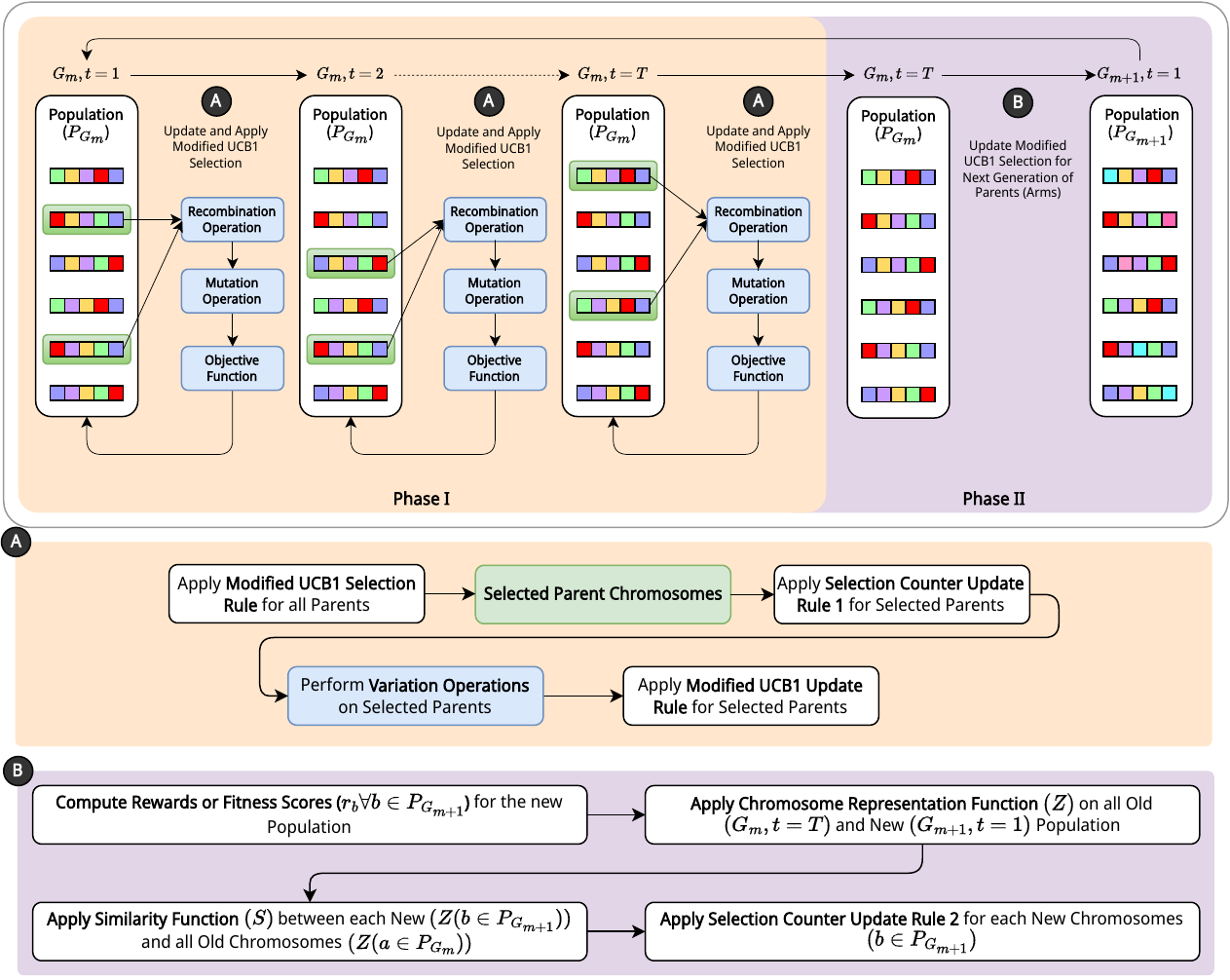}
    \end{center}
    \caption{Illustration of the various steps involved in UBS strategy within simple GA.}
    \label{fig4.2}
\end{figure}

\subsubsection{Phase I}

In Phase I, for a given generation $G_m$, population $P$, and at selection step $t=i$, we choose two parent chromosomes using a modified UCB1-based selection strategy~\eqref{eq4.1}.

\noindent \textit{Modified UCB1 Selection Rule:}
\begin{equation}
    \label{eq4.1}
    A_{t=i} = \underset{S\subset P, |S|=2}{\mathrm{\arg \ \max}} \sum_{a \in S} u_{t=i}^{a}
\end{equation}

Here, $u_{t=i}^{a}$ denotes the modified upper confidence on rewards for parent $a$ at selection step $t=i$, and $|S|=2$ indicates the number of selections we make at a given selection step. Further, once we make the selection, we update the selection counter corresponding to each selected parent using Equation~(\ref{eq4.2}).

\noindent \textit{Selection Counter Update Rule 1:}
\begin{equation}
    \label{eq4.2}
    N_{t=i+1}^{a} \gets N_{t=i}^{a} + 1 \;\forall a \in A_{t=i}.
\end{equation}
Here, $N_{t=i}^a$ denotes the number of times parent $a$ has previously been selected up until $t=i$.

Then, we perform all the variation operations: recombination and mutation. Lastly, we update the modified upper confidence on rewards for parents $a \in A_{t=i}$ at selection step $t = i + 1$ using the modified update rule~\eqref{eq4.7}.

\noindent \textit{Modified UCB1 Update Rule:}
\begin{equation}
    \label{eq4.7}
    u_{t=i+1}^a = r_a + 2 \sqrt{\frac{2 \ln((m\times T) + t)}{N_{t=i+1}^a + 1}} \;\forall a\in A_{t=i}
\end{equation}

{\noindent Here, $r_a$ denotes the fitness score, which is defined in~\eqref{eqfitness}.}

{\begin{equation}
    \label{eqfitness}
r_a = \frac{rank(f_a^{obj}, P)}{|P|}  + (1 - \frac{n_{elite}}{|P|}) \times \frac{\textit{rank}(f_a^{sim}, P)}{|P|}
\end{equation}}

\noindent where $f_a^{obj}$ denotes the objective score of parent a, $n_{elite}$ denotes the number of elite chromosomes to retain, $f_a^{sim}$ denotes the average similarity score of parent $a$ with respect to population $P$. A higher value of $r_a$ is considered better. Therefore, we adjust the ranking order according to the optimization direction, i.e., minimization or maximization. Since $r_a$ does not change within a given generation, we need not recompute them after every selection.

Further, $2\sqrt{2 \frac{\ln((m\times T) + t)}{N^a_{t=i+1}+1}}$ denotes the upper bound around the expected rewards. The Zooming algorithm, which caters to continuum-armed bandit problems when the solution space is Lipschitz continuous \citep{Slivkins2019}, has been the underlying inspiration behind our upper bound approach. The idea behind the upper confidence bound on rewards approach is to use a balanced exploration and exploitation selection strategy where parent chromosomes are selected when rewards are relatively high or when less explored parents exist among the population in a given generation, as informed by the selection counter. We repeat these computations shown in equations~\eqref{eq4.1} through~\eqref{eq4.7} for every selection iteration $t = 1\cdots T$ within each generation $G_m$, where $T$ is the total number of the recombination and mutation (or instead variation) operations.

\subsubsection{Phase II}

Now, at selection step $t = T$, after performing all the computations shown in equations~\eqref{eq4.1} through~\eqref{eq4.7}, we end up having two sets of the population – one containing all the parents from generation $G_m$ and one containing all the offspring generated over the $T$ selection steps along with the elite population from generation $G_m$ (if elitism is applicable) that becomes part of the next generation $G_{m+1}$. {First, we compute the reward or fitness score for the chromosomes in generation $G_{m+1}$.} Further, in Phase II, we apply variation operations, where the parent chromosomes are modified, and the corresponding offspring may or may not share significant overlap with each of the parents. Therefore, creating diversity among the population is essential to prevent premature convergence. However, with the change in the population pool between generations, we have new arms with reward ($r_b$) upper bound starting at the highest value. This change in the population pool creates an artificial pressure to explore all the new arms first. Instead, we propose a similarity-based approach to initialize the upper bound of new arms or (next-generation parent) chromosomes in generation $G_{m+1}$ using the parent chromosomes from previous generation $G_m$. The idea behind a similarity-based approach relies on the belief that parents from the previous generation Gm can collectively help provide a better sense of the search progress for the new chromosomes to continue in generation $G_{m+1}$. Specifically, suppose we have adequately explored parents in generation $G_m$. In that case, we will want to have a smaller upper bound for parents in $G_{m+1}$ that are very similar and likewise a higher upper bound for parents that are highly dissimilar with respect to the parents from the previous generation. However, we do not use a similarity-based approach to update parents' fitness in $G_{m+1}$ since it can be computed deterministically from the objective and similarity between parents in $G_{m+1}$.

The similarity-based approach starts with obtaining a chromosome representation $Z$ in the metric space, depending on the problem. Typically, one hot encoding works for simple optimization problems; however, the representation can become complicated for complex combinatorial optimization problems. This work uses simple chromosome representation schemes such as graph objects or an objective score.

Once we obtain the representation for each of the chromosomes in both the population from $G_m$ and $G_{m+1}$ using $Z(\cdot)$, {we compute the similarity measure of each chromosome in $G_{m+1}$ against each of the chromosomes in $G_m$ using an appropriate similarity function ($Sim$) such as normalized graph edit distance or objective ratios.}

Next, using the similarity measure $Sim$ for each new parent chromosome ($b$) in $G_{m+1}$ ($b \in P_{G_{m+1}}$) against all old parent chromosomes ($a \in P_{G_m}$) and selection counter for old parents $N^a_{t=T}$ from Gm, we initialize the selection counter values $N^b_{t=0}$ for each new parent chromosome ($b$) in $G_{m+1}$ ($b \in P_{G_{m+1}}$) using~\eqref{eq4.8}.

\noindent \textit{Selection Counter Update Rule 2:}

\begin{equation}
    \label{eq4.8}
    N_{t=0}^b = \frac{\sum_{a\in P_G} Sim(a,b)\times N_{t=T}^a }{|P_G|}
\end{equation}

After initializing the selection counter for new chromosomes, Phase II ends, and Phase I starts for generation $G_{m+1}$, and the cycle continues until we meet the termination condition. Note that although all the steps presented here in Phase I apply to generation 0 as well, we intentionally initialize the rewards with normalized objective scores instead of the rank-based fitness function as provided before to encourage early identification of high-quality solutions.

\section{Experiments}\label{experiments}

This section presents the different experimental studies and the setup to evaluate our proposed UBS strategy. Specifically, we evaluate the UBS strategy against several traditional selection approaches on two significantly different NP-hard combinatorial optimization problems: the team orienteering problem (TOP) and the quadratic assignment problem (QAP). Further, for each problem and each selection approach, we first run a characterization experiment to determine the best GA parameter values for population size, recombination probability, and mutation probability. Then, using the best parameter values for each selection approach, we perform a comparison study to determine the effectiveness of the UBS strategy against traditional selection methods. In both the problem scenarios, all the UBS and traditional selection procedures are unchanged. Further, since our goal is not to create a best-in-class GA but to understand the specific benefit of our parent selection strategy, our experiments are tailored accordingly. Nonetheless, if the UBS strategy provides benefits, we could later add it to the best-in-class GA approach to further enhance performance.

\subsection{Setup}

The subsection covers the TOP and QAP problem description, the framework used to evaluate our proposed solution, the different baseline selection approaches used for comparison, the metrics used, the computer software and hardware resources required, and other fixed parameters. Lastly, we cover the problem description, the actual GA configuration, and parameter values used to solve the TOP~\citep{Chao1996} and QAP~\citep{Burkard1997} benchmark problem set.

\subsubsection{Problem Description}

\noindent \textbf{TOP:}

The team orienteering problem (TOP) is a variant of the Orienteering Problem (OP) \citep{Chao1996}. Formally, TOP can be defined as follows. Given a graph $G(V, E)$, where $V$ is a set of vertices, $E$ is a set of edges, the total number of individual/vehicle paths (or plainly, paths) h, travel time for each edge $e \in E : d(e)$, points for each vertex $v \in V : p(v)$ and maximum travel time per path: Tmax, the objective of TOP is to maximize the total points collected from visiting a subset of vertices $P \subseteq V$ collectively forming $h$ paths, each within the maximum travel time per path limit Tmax, such that the starting vertex is $s \in V$ and the ending vertex is $e \in V$ for all the paths.

\noindent \textbf{QAP:}

In QAP \citep{Koopmans1957}, given a set of $n$ locations, $n$ facilities, the distance between each pair of $n$ locations $d_{ij}\forall i, j$ in locations, and flow between each pair of $n$ facilities $f_{ab}$ forall $a, b$ in facilities ($F$), the objective is to find an optimal matching of facilities to the location that minimizes the overall cost involved given by $\sum_{a,b\in F}{f_{ab}\times d_{\phi(a), \phi(b)}}$, where $\phi$ is a feasible solution representing a the permutation matrix that provides a mapping between facility and location indices.

\subsubsection{Evaluation Framework}

We use a subset of TOP and QAP benchmark problems and GA parameters for a given run configuration to execute each selection strategy using an appropriate GA workflow. During a run, we capture the chromosome ID and corresponding objective value for each generation. At the end of a run, we also capture the best objective value found so far. Next, we describe the actual GA configuration/model used to conduct various studies.

\subsubsection{GA Model}

\noindent \textbf{TOP:}

\noindent The underlying GA model for TOP is from \citep{Ferreira2014}.

\noindent \textbf{Representation:} In TOP, we represent the chromosomes as subgraphs of the problem graph $G$, adding edges and nodes that form part of a given solution (or chromosome) to the representation subgraph.

\noindent \textbf{Recombination:} In each application of recombination operation to TOP (selection step $t$), the path corresponding to the vehicle with the highest total score from each parent chromosome (say $a_1$ and $a_2$) becomes the dominant chromosome trait that gets transferred to its offspring. Now, we create the two child chromosomes as follows. First, we create child chromosomes $b_1$ and $b_2$ by copying all elements from $a_2$ and $a_1$, respectively, and then perform the following modifications. Add chromosome $a_1$'s best scoring path (say $a^{best}_1$) to $b_1$ and then delete all duplicate vertices found in $b_1$ that are already in $a^{best}_1$ (equivalent to $b_1 = a^{best}_1 \cup \{a_2 - a^{best}_1 \}$). Now, delete the path with the total lowest score from $b_1$ to ensure the constraint on the number of vehicles ($|A|$) is satisfied. Similarly, $b_2 = a^{best}_2 \cup \{a_1 - a^{best}_2\}$.

\noindent \textbf{Mutation:} Given a chromosome, we randomly delete a vertex from a random path. Then, we recursively attempt to add as many unused vertices as possible at a random position to each path (in random order) until no further additions are feasible. Note that we only attempt to add each of the remaining vertices once.

\noindent \textbf{Elite Subpopulation:} 10\% of the total population.

\noindent \textbf{QAP:}

\noindent The underlying GA model for QAP is from \citep{Azarbonyad2014}.

\noindent \textbf{Representation:} Permutation vector

\noindent \textbf{Recombination:} Two-point crossover operation.

\noindent \textbf{Mutation:} Following \citep{Azarbonyad2014}, we randomly swap the permutation vector for mutation.

\noindent \textbf{Elite Subpopulation:} 10\% of the total population.

\subsubsection{Traditional Selection Strategies for Comparison}

This section briefly provides the different traditional selection strategies with two variants used for comparison. For additional reading on the various traditional selection methods used in this work, we recommend readers to the comparative studies by \citep{Blickle1996,Jebari2013,Shukla2015} on the various selection methods and review by \citep{Katoch2021} on genetic algorithms. For traditional selection strategies, except for uniform random selection, all use an intermediate population. The general description of the different baseline selection methods used is as follows.

\begin{enumerate}
    \item[a.] \textbf{Uniform Random Selection (URS):} In this selection strategy, for a given generation and selection step, the two parent chromosomes are selected at random using a uniform distribution without considering the fitness (or objective) scores. URS is a baseline selection strategy used to understand the difficulty of a problem to search for high-quality solutions.
    
    \item[b.] \textbf{Roulette Wheel Selection (RWS):} RWS is a variant of the proportionate selection and starts with sectors on a roulette wheel divided among the different chromosomes proportional to their fitness. Now, we randomly select a starting position for the pointer and spin the wheel - the chromosome corresponding to the pointer location when the wheel stops are selected. The process repeats until the limit on the total number of selections is satisfied.
    \item[c.] \textbf{Ranking Selection (RS):} In the RS strategy, we first assign ranks to all the chromosomes based on their fitness, e.g., the largest rank for the chromosome with the best fitness. Then, using this rank instead of fitness, we perform selection with a proportionate selection strategy like RWS. The process repeats until the limit on the total number of selections is satisfied.
    \item[d.] \textbf{Tournament Selection (TS):} In this strategy, we randomly select two chromosomes, compare their fitness values, and select the one with the best fitness. The process repeats until the limit on the total number of selections is satisfied. Further, since the rank-based fitness function used as rewards in the UBS strategy was proposed initially for use with a Tournament selection \citep{Vidal2022}, we also use it as the fitness function for the baseline TS strategy.
    \item[e.] \textbf{Stochastic Universal Sampling (SUS):} SUS is a variant of proportionate selection that aims to alleviate the inherent selection bias of independent sampling. The SUS strategy starts with sectors on a roulette wheel divided among the different chromosomes proportional to their reproduction rate given as a ratio of fitness to average population fitness. Instead of a single pointer, we use N equidistant pointers equal to the total number of selections. Next, we randomly select a starting position for one of the pointers and spin the wheel. The chromosomes corresponding to each pointer location when the wheel stops are all selected in one go. The process repeats until the limit on the total number of selections is satisfied.
\end{enumerate}

\subsubsection{Upper Bound-based Parent Select (UBS)}

For UBS, we follow the selection methodology provided in section~\ref{UBS}. We only modify three aspects to address the problem instances: TOP and QAP, which are significantly different. Firstly, the fitness function is modified to incorporate the sense of the optimization direction, i.e., maximization for TOP and minimization for QAP. Secondly, the reward initialization is modified to ensure that one corresponds to the best value and a proportionally lower value to others. Thirdly, the similarity function used for TOP is a variant of the graph edit distance \citep{Sanfeliu1983} based similarity, which compares two labeled graph inputs for the similarity between undirected edges and nodes, while for QAP, we use the ratio of the objective values.

\subsubsection{Test Dataset}

\noindent \textbf{TOP:} The TOP benchmark problem dataset \citep{Chao1996} contains 49 feasible large problems with 100 to 102 vertices, 2 to 4 paths, and varying maximum travel length restrictions (Tmax). We divide the dataset into two parts: 1/3rd ($\approx 19$) reserved for characterization experiments, and the remaining 2/3rd ($\approx 30$) for comparison experiments. The problem instances in each experimental category are randomly decided and fixed throughout the study.

\noindent \textbf{QAP:} The QAP benchmark problem dataset \citep{Burkard1997} contains 61 feasible problems ranging from 30 through 256 location-facility pairs with symmetric or asymmetric flow and travel distance costs. Similar to TOP, we divide the dataset into two parts: 1/3rd ($\approx 20$) reserved for characterization experiments, and the remaining 2/3rd ($\approx 41$) for comparison experiments. The problem instances in each experimental category are randomly decided and fixed throughout the study.

\subsubsection{Characterization Study Specification}

Each characterization experiment on the problem instances from a benchmark dataset with a given selection method is run with a time limit of 5 minutes and for three replications with different random seeds. Further, three parameters are varied for each run: population size, recombination probability, and mutation probability. We used four different configurations for population size: 50,100,150, and 200. Further, we use ten different configurations for recombination probabilities: 0.1 through 1.0 with 0.1 increments. Lastly, we use four mutation probabilities: 0.001, 0.01, 0.1, and 1.0. The intuition behind using the different values for the recombination and mutation probabilities stems from the evidence found in the literature on the use of larger values for recombination usually provided better results while a lower value of mutation for the model under GA consideration \citep{Azarbonyad2014,Ferreira2014}. However, the effect of these parameters is unknown with respect to UBS. Hence, we used a logarithmic step size for the mutation parameter to cover a larger area of the hyperparameter space.

\subsubsection{Comparison Study Specification}

Each comparison experiment on the problem instances from a benchmark dataset with a given selection method with the best GA configuration found from the previous study is run with a time limit of 5 minutes and for ten replications with different random seeds.

\subsubsection{Performance Measure}

In the characterization experiment, we use the average relative percentage error ($ARPE$) following \citep{Dang2013}, defined as $abs(\frac{1}{|RS|}\underset{j\in RS}{\sum} Obj_{ij}^k - BKS_i) \times \frac{100}{BKS_i}$, where $Obj_{ij}^k$ is the best objective function value found by the GA using a given selection strategy $k$, random seed $j$ (from the set of random seeds: $RS$) and configuration while $BKS_i$ is the best-known solution found from the literature for problem instance $i$. Further, in the comparison experiments, apart from $ARPE$, we also report relative percentage error ($RPE$) and median relative percentage error ($MRPE$) inspired by \citep{Dang2013}. Here, RPE and MRPE are defined as $abs(\frac{1}{|RS|}\underset{j\in RS}{\max}\ Obj_{ij}^k - BKS_i)\times \frac{100}{BKS_i}$ and $abs(\frac{1}{|RS|} \underset{j\in RS}{median}\ Obj_{ij}^k - BKS_i) \times \frac{100}{BKS_i}$, respectively.

\subsubsection{Computing Resources}

All studies use a computing cluster (Rochester Institute of Technology, 2019). Each problem instance uses 1 GB RAM, one core, and two logical threads of Intel Xeon Gold 6150 CPU $@$ 2.70GHz. Further, the entire genetic algorithm and various selection strategies are all coded in Python without parallelization.

\section{Results}

In this section, we will first present the results from the characterization and comparison studies for TOP, followed by the characterization and comparison studies for QAP.

\subsection{Characterization Study on TOP}\label{Sec-TOP-Charaterization}

Figures~\ref{fig6.1} through~\ref{fig6.6} show the performance plots of different selection mechanisms across different GA hyperparameters on the TOP benchmark dataset. Within each plot, we can see multiple line subplots showing the performance of the GA algorithm under a given selection strategy and configuration. The configuration reads as follows for a given point in the plot. The x-axis reads the recombination probability used, the secondary y-axis shows the selection strategy used, and two nested titles for each subplot are on the top (secondary x-axis). At the top level, we have the mutation probability; at the bottom, we have the population size. Further, with each subplot, two points, the highest and lowest values, are annotated with their corresponding values.

From Figures~\ref{fig6.1} through~\ref{fig6.6}, we notice that as we move from a lower mutation probability to a higher one, there is consistent improvement in performance across all the selection methods. Further, we achieve the best results when the mutation probability is 1.

Next, in Figure~\ref{fig6.1}, which shows the characterization performance of the proposed UBS strategy, the interesting aspect is that for low mutation probability, the performance is relatively weak, irrespective of population size. However, increasing the mutation probability improves performance across different population sizes while holding recombination probabilities constant. Therefore, it is evident that large changes between generations are mostly favorable to the UBS strategy. Also, the effect of population size is more pronounced when mutation probabilities are above 0.1. However, in contrast, all other selection strategies (figures~\ref{fig6.2} through~\ref{fig6.6}) show improvement when population size increases, irrespective of the mutation probability used.

Further, looking at Figures~\ref{fig6.1} through~\ref{fig6.6} with respect to population size, we gather that we find the best results when the population size is 200.

Lastly, across all the figures, it is observed that the performance generally improves on the TOP instances when increasing recombination probabilities while holding other parameters constant. Table~\ref{tab:top-best-config} shows the best configuration obtained from the characterization experiment for various selection strategies on TOP.

\begin{figure}[H]
    \begin{center}
        \includegraphics[width=1\linewidth]{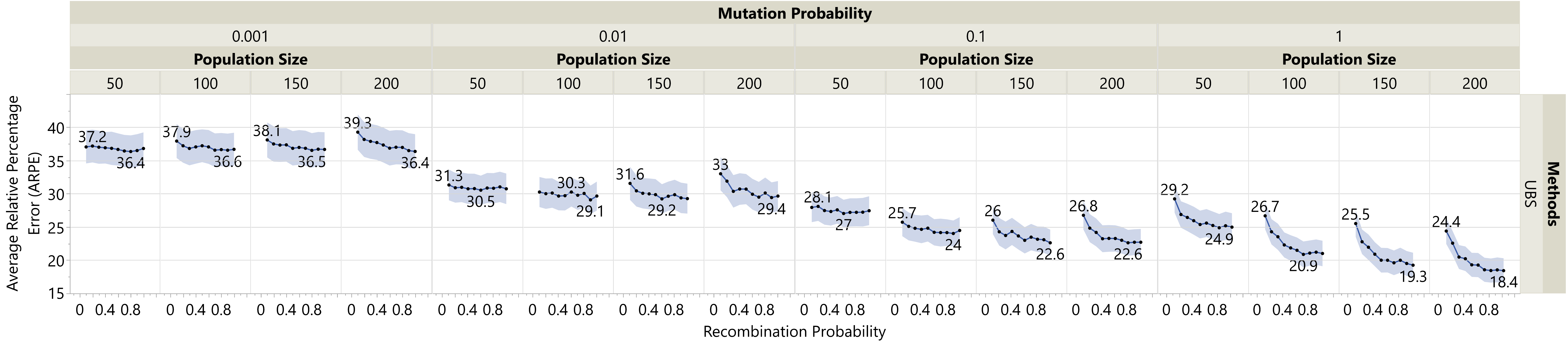}
    \end{center}
    \caption{Characterization of UBS at different population sizes, recombination probabilities, and mutation probabilities. The points (black) denote the mean(measure) across different TOP problem instances. The measure used is the Average Relative Percentage Error, and the band (blue) denotes the standard error about the mean. The values annotated are the highest and lowest values within each line plot. Lower APRE values are desirable.}
    \label{fig6.1}
\end{figure}

\begin{figure}[H]
    \begin{center}
        \includegraphics[width=1\linewidth]{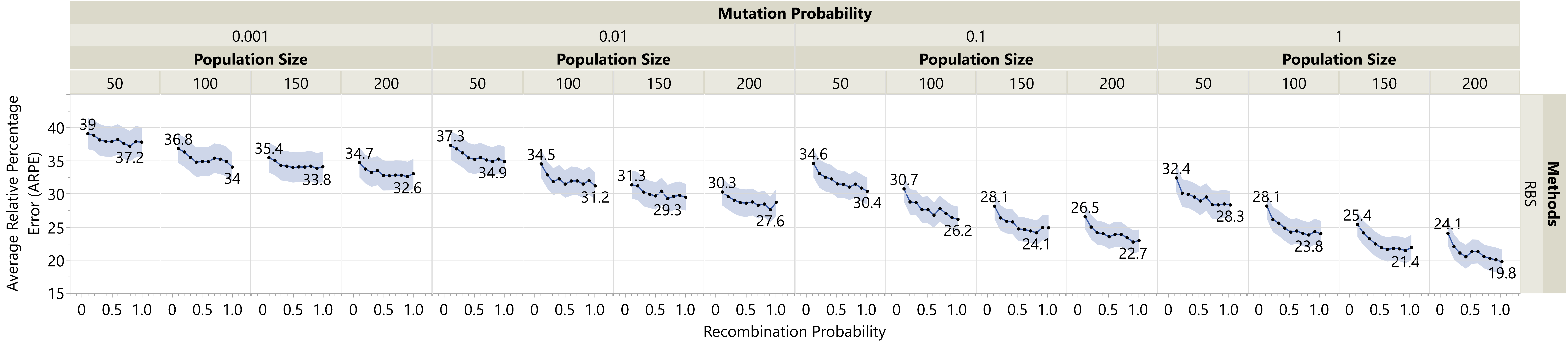}
    \end{center}
    \caption{Characterization of RBS at different population sizes, recombination probabilities, and mutation probabilities. The points (black) denote the mean(measure) across different TOP problem instances. The measure used is the Average Relative Percentage Error, and the band (blue) denotes the standard error about the mean. The values annotated are the highest and lowest values within each line plot. Lower APRE values are desirable.}
    \label{fig6.2}
\end{figure}

\begin{figure}[H]
    \begin{center}
        \includegraphics[width=1\linewidth]{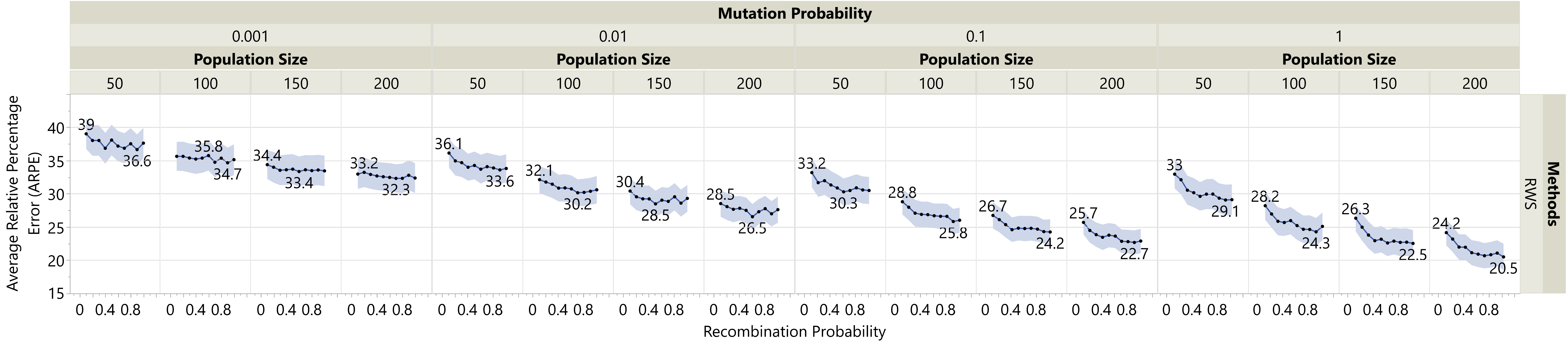}
    \end{center}
    \caption{Characterization of RWS at different population sizes, recombination probabilities, and mutation probabilities. The points (black) denote the mean(measure) across different TOP problem instances. The measure used is the Average Relative Percentage Error, and the band (blue) denotes the standard error about the mean. The values annotated are the highest and lowest values within each line plot. Lower APRE values are desirable.}
    \label{fig6.3}
\end{figure}

\begin{figure}[H]
    \begin{center}
        \includegraphics[width=1\linewidth]{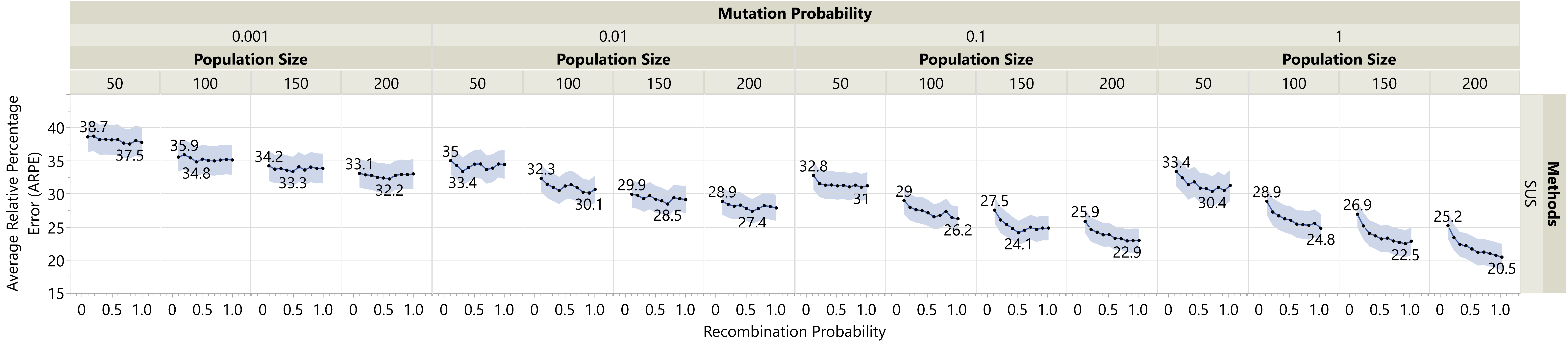}
    \end{center}
    \caption{Characterization of SUS at different population sizes, recombination probabilities, and mutation probabilities. The points (black) denote the mean(measure) across different TOP problem instances. The measure used is the Average Relative Percentage Error, and the band (blue) denotes the standard error about the mean. The values annotated are the highest and lowest values within each line plot. Lower APRE values are desirable.}
    \label{fig6.4}
\end{figure}

\begin{figure}[H]
    \begin{center}
        \includegraphics[width=1\linewidth]{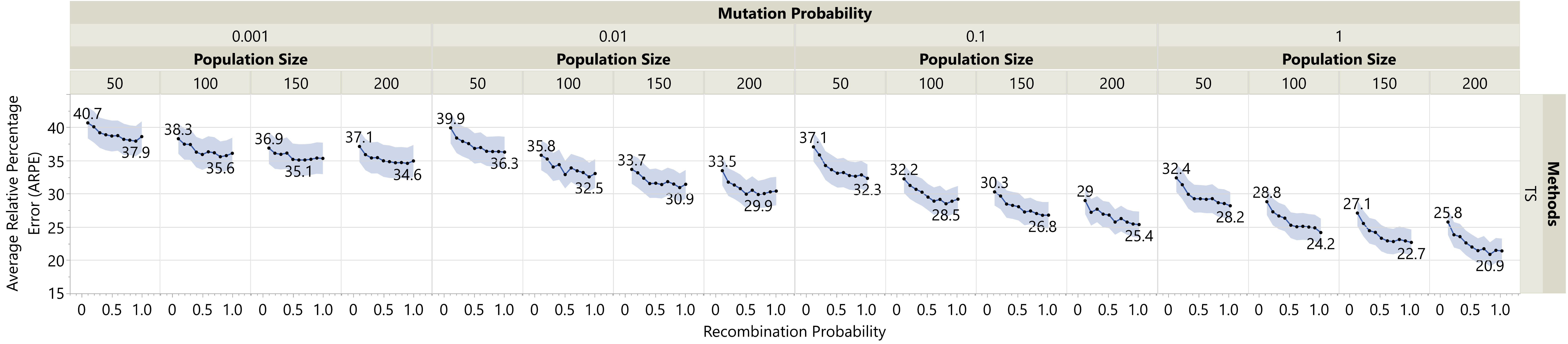}
    \end{center}
    \caption{Characterization of TS at different population sizes, recombination probabilities, and mutation probabilities. The points (black) denote the mean(measure) across different TOP problem instances. The measure used is the Average Relative Percentage Error, and the band (blue) denotes the standard error about the mean. The values annotated are the highest and lowest values within each line plot. Lower APRE values are desirable.}
    \label{fig6.5}
\end{figure}

\begin{figure}[H]
    \begin{center}
        \includegraphics[width=1\linewidth]{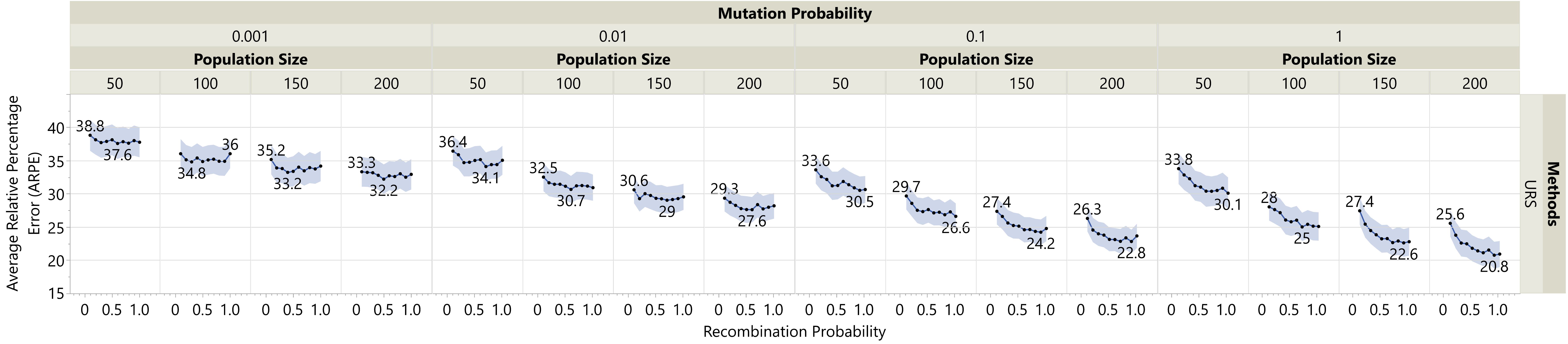}
    \end{center}
    \caption{Characterization of URS at different population sizes, recombination probabilities, and mutation probabilities. The points (black) denote the mean(measure) across different TOP problem instances. The measure used is the Average Relative Percentage Error, and the band (blue) denotes the standard error about the mean. The values annotated are the highest and lowest values within each line plot. Lower APRE values are desirable.}
    \label{fig6.6}
\end{figure}

\begin{table}[H]
    \caption{Best configuration obtained from the characterization experiment for various selection strategies on TOP}
    \label{tab:top-best-config}
    \centering
    \resizebox{0.71\columnwidth}{!}{
    \begin{tabular*}{\columnwidth}{C{0.125\columnwidth} C{0.25\columnwidth} C{0.25\columnwidth} M{0.25\columnwidth}}
    \toprule
    \multicolumn{1}{c}{Methods} & \multicolumn{1}{c}{Population Size} & \multicolumn{1}{c}{Recombination Probability} & Mutation Probability \\ \midrule
    UBS & 200 & 0.8 & 1.0 \\
    RBS & 200 & 0.8 & 1.0 \\
    TS  & 200 & 0.8 & 1.0 \\
    SUS & 200 & 0.9 & 1.0 \\
    URS & 200 & 0.7 & 1.0 \\
    RWS & 200 & 0.9 & 1.0 \\ \bottomrule
    \end{tabular*}}
    \end{table}

\subsection{Comparison Study on TOP}

The results from experimental runs are analyzed using a non-parametric Wilcoxon-Signed Rank matched pairwise comparison test~\citep{Rey2011} at a significant level of 0.05, available within the JMP software package~\citep{SAS2020}.
Table~\ref{tab:top} shows that UBS outperforms all the other traditional strategies across all performance measures. Interestingly, although TS and UBS strategies share the same fitness function, the observed performance is significantly different. We believe this is partly due to the greedy upper bound-based selection, i.e., considering all existing populations and knowledge transfer between generations within the UBS strategy.

\begin{table}[H]
    \caption{Results from the comparison experiments on TOP. The best result that is statistically significant is bolded.}
    \label{tab:top}
    \centering
    \resizebox{0.85\linewidth}{!}{
    \begin{tabular*}{1.15\linewidth}{@{}cccc@{}}
    \toprule
    Methods & Average Relative Percentage Error   (ARPE) & Median Relative Percentage Error   (ARPE) & Relative Percentage Error (RPE) \\ \midrule
    \textbf{UBS} & \textbf{18.84} & \textbf{18.71} & \textbf{14.53} \\
    RBS & 20.28 & 20.16 & 16.14 \\
    RWS & 21.20 & 21.02 & 17.12 \\
    TS  & 21.22 & 21.11 & 16.63 \\
    SUS & 21.33 & 21.48 & 16.45 \\
    URS & 21.37 & 21.39 & 16.56 \\ \bottomrule
    \end{tabular*}}
    \end{table}

\subsection{Characterization Study on QAP}

The general description of how to read the plots shown in Figures~\ref{fig6.7} through~\ref{fig6.12} is the same as in subsection~\ref{Sec-TOP-Charaterization}. Here, we will provide our experimental observations.

Figures~\ref{fig6.7} through~\ref{fig6.12} show that the mutation and recombination probability provide an interesting interaction effect. In general, we see that as we increase the mutation probability, the performance improves when the recombination probability is low ($\sim$ 0.1). However, we observe an exception in the case of UBS. In UBS, the performance improves across the different values of recombination probability, regardless. This aspect reinforces our earlier observations on the favoritism shown by UBS toward large changes between generations.

Next, in figures~\ref{fig6.8} through~\ref{fig6.12}, we notice that increasing both mutation and recombination probability has a detrimental effect on all the traditional selection strategies. Interestingly, in Figure~\ref{fig6.7}, UBS showed a more robust performance under a similar observational setting.

Lastly, across all the figures, we gather that the best configuration in general for all the selection strategies has the highest mutation probability of 1.0 with varied population size and recombination probability. Refer to Table~\ref{tab:qap-best-config}, which shows the best configuration obtained from the characterization experiment for various selection strategies on QAP.

\begin{figure}[H]
    \begin{center}
        \includegraphics[width=1\linewidth]{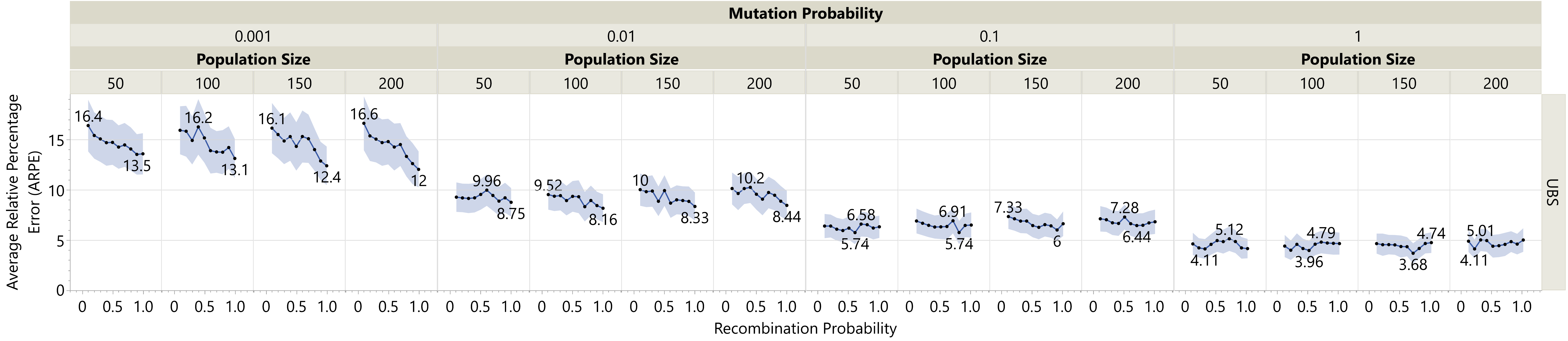}
    \end{center}
    \caption{Characterization of UBS at different population sizes, recombination probabilities, and mutation probabilities. The points (black) denote the mean(measure) across different QAP problem instances. The measure used is the Average Relative Percentage Error, and the band (blue) denotes the standard error about the mean. The values annotated are the highest and lowest values within each line plot. Lower APRE values are desirable.}
    \label{fig6.7}
\end{figure}

\begin{figure}[H]
    \begin{center}
        \includegraphics[width=1\linewidth]{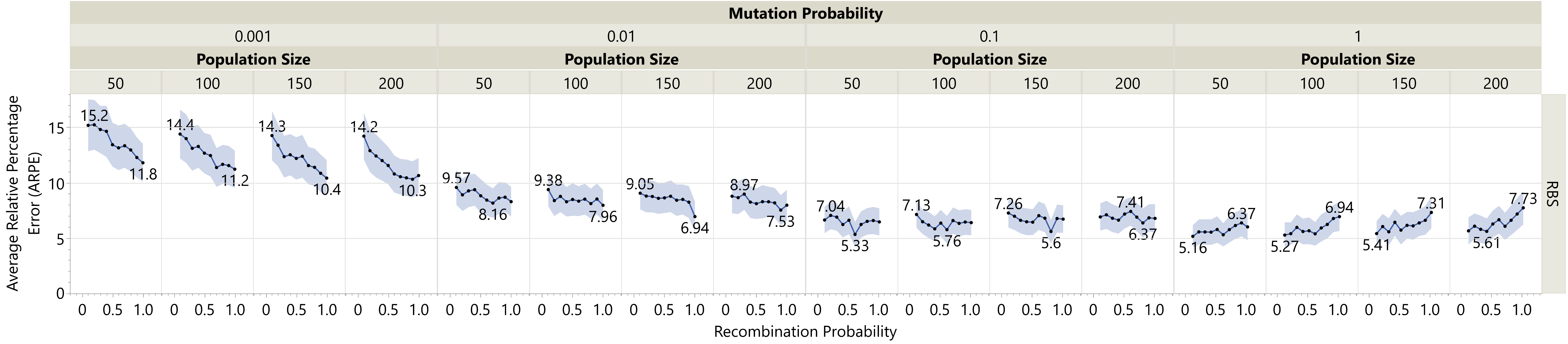}
    \end{center}
    \caption{Characterization of RBS at different population sizes, recombination probabilities, and mutation probabilities. The points (black) denote the mean(measure) across different QAP problem instances. The measure used is the Average Relative Percentage Error, and the band (blue) denotes the standard error about the mean. The values annotated are the highest and lowest values within each line plot. Lower APRE values are desirable.}
    \label{fig6.8}
\end{figure}

\begin{figure}[H]
    \begin{center}
        \includegraphics[width=1\linewidth]{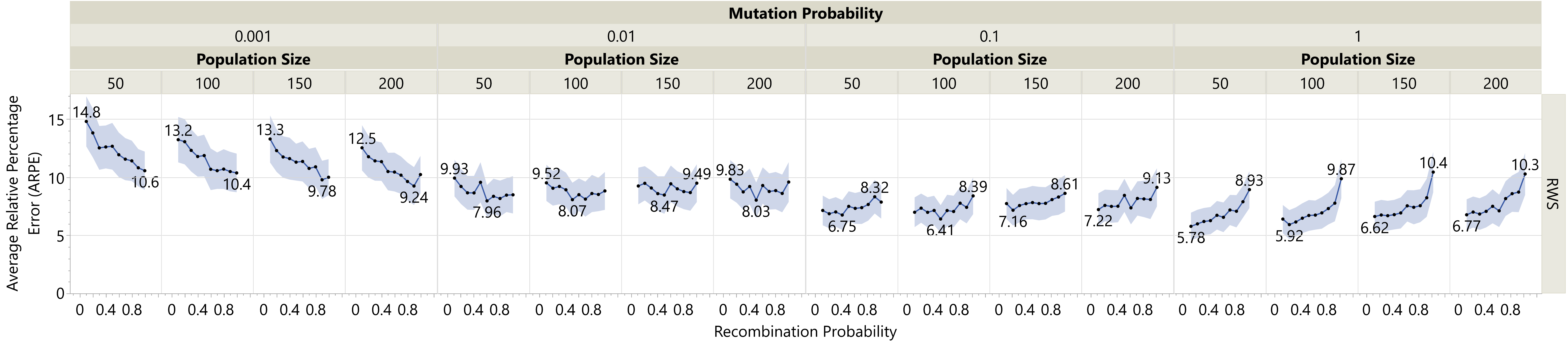}
    \end{center}
    \caption{Characterization of RWS at different population sizes, recombination probabilities, and mutation probabilities. The points (black) denote the mean(measure) across different QAP problem instances. The measure used is the Average Relative Percentage Error, and the band (blue) denotes the standard error about the mean. The values annotated are the highest and lowest values within each line plot. Lower APRE values are desirable.}
    \label{fig6.9}
\end{figure}

\begin{figure}[H]
    \begin{center}
        \includegraphics[width=1\linewidth]{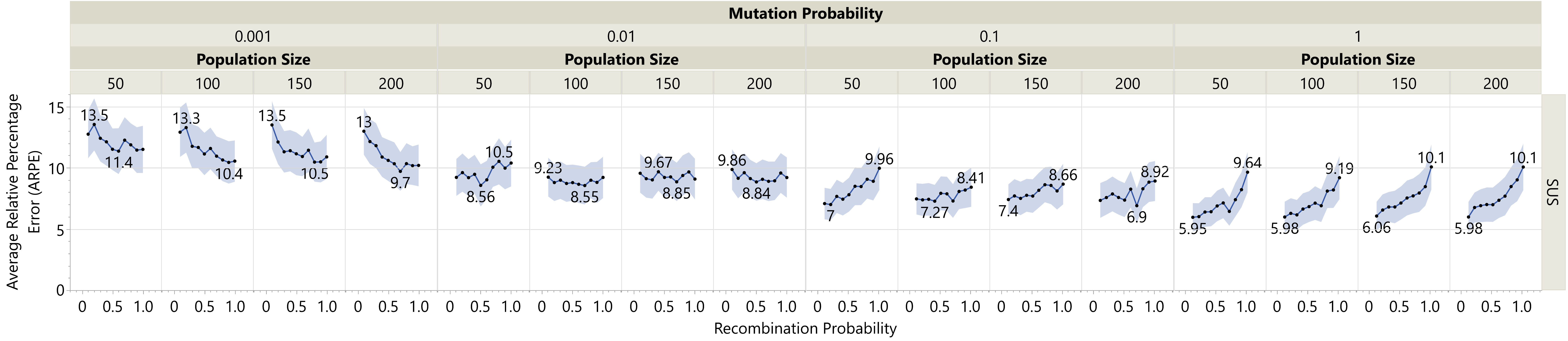}
    \end{center}
    \caption{Characterization of SUS at different population sizes, recombination probabilities, and mutation probabilities. The points (black) denote the mean(measure) across different QAP problem instances. The measure used is the Average Relative Percentage Error, and the band (blue) denotes the standard error about the mean. The values annotated are the highest and lowest values within each line plot. Lower APRE values are desirable.}
    \label{fig6.10}
\end{figure}

\begin{figure}[H]
    \begin{center}
        \includegraphics[width=1\linewidth]{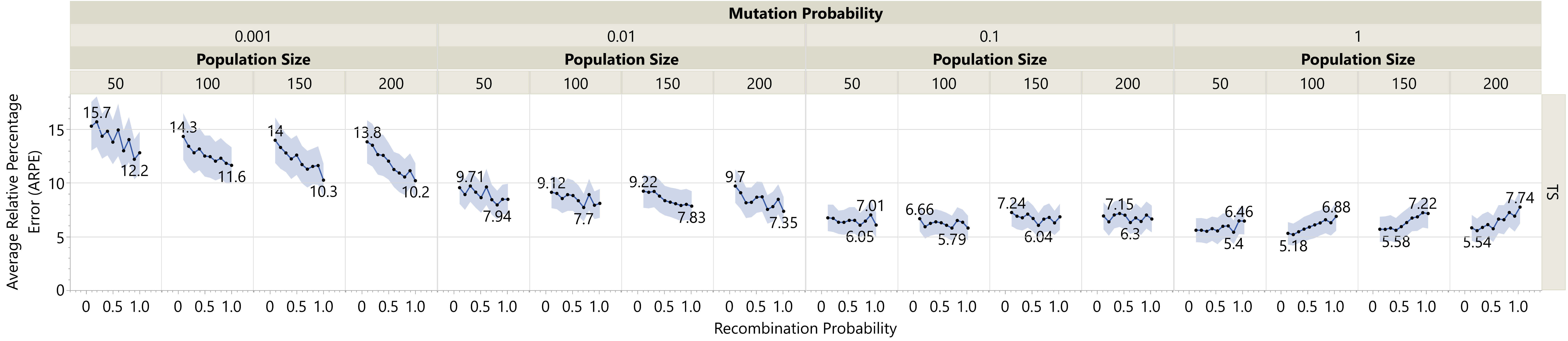}
    \end{center}
    \caption{Characterization of TS at different population sizes, recombination probabilities, and mutation probabilities. The points (black) denote the mean(measure) across different QAP problem instances. The measure used is the Average Relative Percentage Error, and the band (blue) denotes the standard error about the mean. The values annotated are the highest and lowest values within each line plot. Lower APRE values are desirable.}
    \label{fig6.11}
\end{figure}

\begin{figure}[H]
    \begin{center}
        \includegraphics[width=1\linewidth]{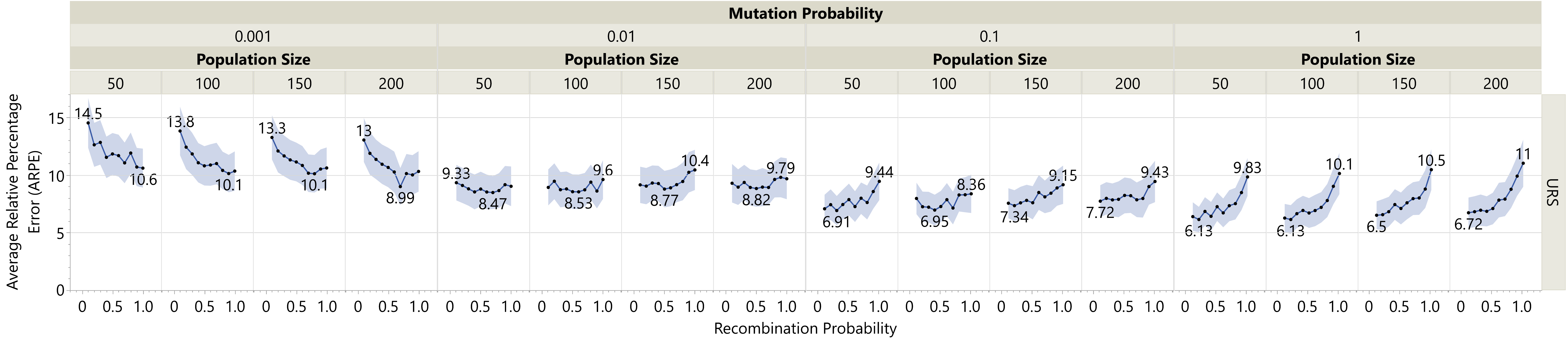}
    \end{center}
    \caption{Characterization of URS at different population sizes, recombination probabilities, and mutation probabilities. The points (black) denote the mean(measure) across different QAP problem instances. The measure used is the Average Relative Percentage Error, and the band (blue) denotes the standard error about the mean. The values annotated are the highest and lowest values within each line plot. Lower APRE values are desirable.}
    \label{fig6.12}
\end{figure}

\begin{table}[H]
    \caption{Best configuration obtained from the characterization experiment for various selection strategies on QAP}
    \label{tab:qap-best-config}
    \centering
    \resizebox{0.71\columnwidth}{!}{
    \begin{tabular*}{\columnwidth}{C{0.125\columnwidth} C{0.25\columnwidth} C{0.25\columnwidth} M{0.25\columnwidth}}
    \toprule
    \multicolumn{1}{c}{Methods} & \multicolumn{1}{c}{Population Size} & \multicolumn{1}{c}{Recombination Probability} & Mutation Probability \\ \midrule
    UBS & 150 & 0.7 & 1.0 \\
    RBS & 50  & 0.1 & 1.0 \\
    TS  & 100 & 0.2 & 1.0 \\
    RWS & 50  & 0.1 & 1.0 \\
    SUS & 50  & 0.1 & 1.0 \\
    URS & 100 & 0.2 & 1.0 \\ \bottomrule
    \end{tabular*}}
    \end{table}

\subsection{Comparison Study on QAP}

The results from experimental runs are analyzed using a non-parametric Wilcoxon-Signed Rank matched pairwise comparison test~\citep{Rey2011} at a significant level of 0.05, available within the JMP software package~\citep{SAS2020}.

Table~\ref{tab:qap} shows that UBS again outperforms all the other traditional strategies across all performance measures. We observe a similar performance difference between the TS and UBS strategy, which shares the same fitness function, reinforcing our earlier observation on the effectiveness of the greedy upper bound-based selection and knowledge transfer between generations within the UBS strategy.

\begin{table}[H]
    \caption{Results from the comparison experiments on QAP. The best result that is statistically significant is bolded.}
    \label{tab:qap}
    \centering
    \resizebox{0.85\linewidth}{!}{
    \begin{tabular*}{1.15\linewidth}{@{}cccc@{}}
    \toprule
    Methods & Average Relative Percentage Error   (ARPE) & Median Relative Percentage Error (ARPE) & Relative Percentage Error (RPE) \\ \midrule
    \textbf{UBS}	& \textbf{6.15}	& \textbf{6.21}	& \textbf{4.32} \\
    RBS	& 6.70	& 6.76	& 4.84 \\
    RWS	& 7.03	& 7.10	& 5.47 \\
    TS	& 7.61	& 7.58	& 5.59 \\
    SUS	& 7.64	& 7.62	& 5.34 \\
    URS	& 8.06	& 7.95	& 6.32 \\ \bottomrule    
    \end{tabular*}}
    \end{table}

\section{Conclusions}

In this work, we proposed a deterministic parent selection strategy for recombination called Upper Bound-based Parent Selection (UBS) in a generational GA setting to solve NP-hard combinatorial optimization problems. We formulated the parent selection problem using the MAB framework as part of the UBS strategy. Then, we solved it using a modified UCB1 algorithm. Specifically, the modified UCB1 algorithm used a combination of a ranked-based fitness function that preserves diversity as rewards and an upper-bound approach to managing exploration and exploitation. Further, we provided a unique similarity-based approach for transferring knowledge of the search progress between generations to accelerate the search.

Next, we demonstrated the effectiveness of the proposed UBS strategy in comparison to traditional stochastic selection strategies through extensive experiments on two NP-hard combinatorial optimization problem instances: the team orienteering and the quadratic assignment benchmark problems. The results from the characterization studies revealed that UBS, in most cases, favors large variations among the population between generations. Next, results from the comparison studies revealed that UBS can effectively search for high-quality solutions faster than traditional stochastic selection strategies on challenging NP-hard combinatorial optimization problems.

\section{Limitations and Future Work}

Some of the current limitations of UBS include the following. UBS is sensitive to the variation operators - recombination and mutation, as can be noticed from the characterization experiments. Also, during our initial screening experiments, we observed that the UBS approach is sensitive to the choice of the similarity measure used. We aim to perform more characteristic experiments on different similarity measures in the future. Further, we plan to integrate the UBS approach with a state-of-the-art implementation to understand the benefits of such approaches on sophisticated evolutionary algorithms. Lastly, we would like to study the effectiveness of the UBS algorithm with multiple parent recombination (more than two) in a steady-state GA setting.

\section*{Acknowledgment}

This work was partly supported by the Office of Naval Research under Contract No. N00014-20-1-2242. Further, the authors acknowledge RIT Research Computing \citep{RochesterInstituteofTechnology2019} for providing the necessary computing resources and timely support.



\small
\bibliographystyle{unsrtnat}
\bibliography{UBS}

\begin{thebibliography}{40}
\providecommand{\natexlab}[1]{#1}
\providecommand{\url}[1]{\texttt{#1}}
\expandafter\ifx\csname urlstyle\endcsname\relax
  \providecommand{\doi}[1]{doi: #1}\else
  \providecommand{\doi}{doi: \begingroup \urlstyle{rm}\Url}\fi

\bibitem[Talbi(2009)]{Talbi2009}
{El-ghazali} Talbi.
\newblock \emph{Metaheuristics: From Design to Implementation}.
\newblock John Wiley and Sons, Incorporated, 1 edition, 2009.
\newblock ISBN 978-0-470-49690-9.

\bibitem[Ghaheri et~al.(2015)Ghaheri, Shoar, Naderan, and Hoseini]{Ghaheri2015}
Ali Ghaheri, Saeed Shoar, Mohammad Naderan, and Sayed~Shahabuddin Hoseini.
\newblock The applications of genetic algorithms in medicine.
\newblock \emph{Oman Medical Journal}, 30:\penalty0 406--416, 2015.
\newblock ISSN 20705204.
\newblock \doi{10.5001/omj.2015.82}.

\bibitem[Yang(2014)]{Yang2014}
Xin-She Yang.
\newblock \emph{Genetic Algorithms}.
\newblock Elsevier, 2014.
\newblock \doi{10.1016/B978-0-12-416743-8.00005-1}.

\bibitem[Sivanandam and Deepa(2008)]{Sivanandam2008}
S~N Sivanandam and S~N Deepa.
\newblock \emph{Introduction to Genetic Algorithms}.
\newblock Springer Berlin Heidelberg, 2008.
\newblock ISBN 978-3-540-73189-4.
\newblock \doi{10.1007/978-3-540-73190-0}.
\newblock URL \url{http://link.springer.com/10.1007/978-3-540-73190-0}.

\bibitem[Back(1994)]{Back1994}
T.~Back.
\newblock Selective pressure in evolutionary algorithms: a characterization of selection mechanisms.
\newblock In \emph{Proceedings of the First IEEE Conference on Evolutionary Computation. IEEE World Congress on Computational Intelligence}, pages 57--62. IEEE, 1994.
\newblock ISBN 0-7803-1899-4.
\newblock \doi{10.1109/ICEC.1994.350042}.
\newblock URL \url{http://ieeexplore.ieee.org/document/350042/}.

\bibitem[Blickle and Thiele(1996)]{Blickle1996}
Tobias Blickle and Lothar Thiele.
\newblock A comparison of selection schemes used in evolutionary algorithms.
\newblock \emph{Evolutionary Computation}, 4:\penalty0 361--394, 12 1996.
\newblock ISSN 1063-6560.
\newblock \doi{10.1162/evco.1996.4.4.361}.
\newblock URL \url{https://direct.mit.edu/evco/article/4/4/361-394/769}.

\bibitem[Crepinsek et~al.(2013)Crepinsek, Liu, and Mernik]{Crepinsek2013}
Matej Crepinsek, Shih~Hsi Liu, and Marjan Mernik.
\newblock Exploration and exploitation in evolutionary algorithms: A survey.
\newblock \emph{ACM Computing Surveys}, 45:\penalty0 1--33, 6 2013.
\newblock ISSN 03600300.
\newblock \doi{10.1145/2480741.2480752}.

\bibitem[Rogers and Prugel-Bennett(1999)]{Rogers1999}
Alex Rogers and Adam Prugel-Bennett.
\newblock Genetic drift in genetic algorithm selection schemes.
\newblock \emph{IEEE Transactions on Evolutionary Computation}, 3:\penalty0 298--303, 1999.
\newblock ISSN 1089778X.
\newblock \doi{10.1109/4235.797972}.

\bibitem[Reeves(2010)]{Reeves2010}
Colin~R. Reeves.
\newblock \emph{Genetic Algorithms}, volume 146.
\newblock Springer US, 2010.
\newblock \doi{10.1007/978-1-4419-1665-5\_5}.

\bibitem[Srinivas and Patnaik(1994)]{Srinivas1994}
M.~Srinivas and Lalit~M. Patnaik.
\newblock Genetic algorithms: A survey.
\newblock \emph{Computer}, 27:\penalty0 17--26, 1994.
\newblock ISSN 00189162.
\newblock \doi{10.1109/2.294849}.

\bibitem[Chao et~al.(1996)Chao, Golden, and Wasil]{Chao1996}
{I-Ming} Chao, {Bruce L.} Golden, and {Edward a.} Wasil.
\newblock The team orienteering problem.
\newblock \emph{European Journal of Operational Research}, 88:\penalty0 464--474, 2 1996.
\newblock ISSN 03772217.
\newblock \doi{10.1016/0377-2217(94)00289-4}.
\newblock URL \url{https://linkinghub.elsevier.com/retrieve/pii/0377221794002894}.

\bibitem[Koopmans and Beckmann(1957)]{Koopmans1957}
Tjalling~C. Koopmans and Martin Beckmann.
\newblock Assignment problems and the location of economic activities.
\newblock \emph{Econometrica}, 25:\penalty0 53, 1 1957.
\newblock ISSN 00129682.
\newblock \doi{10.2307/1907742}.
\newblock URL \url{https://www.jstor.org/stable/1907742?origin=crossref}.

\bibitem[Auer(2002)]{Auer2002}
Peter Auer.
\newblock Using confidence bounds for exploitation-exploration trade-offs.
\newblock \emph{Journal of Machine Learning Research}, 3:\penalty0 397--422, 2002.

\bibitem[Auer and Fischer(2002)]{AuerFischer2002}
Peter Auer and Paul Fischer.
\newblock Finite-time analysis of the multiarmed bandit problem.
\newblock \emph{Machine Learning}, 47:\penalty0 235--256, 2002.
\newblock \doi{https://doi.org/10.1023/A:1013689704352}.
\newblock URL \url{https://link.springer.com/article/10.1023/A:1013689704352}.

\bibitem[Slivkins(2019)]{Slivkins2019}
Aleksandrs Slivkins.
\newblock Introduction to multi-armed bandits.
\newblock \emph{Foundations and Trends® in Machine Learning}, 12:\penalty0 1--286, 4 2019.
\newblock ISSN 1935-8237.
\newblock \doi{10.1561/2200000068}.
\newblock URL \url{http://www.nowpublishers.com/article/Details/MAL-068}.

\bibitem[Sutton and Barto(2018)]{Sutton2018}
Richard~S. Sutton and Andrew~G. Barto.
\newblock \emph{Reinforcement Learning: An Introduction}.
\newblock MIT Press, second edition, 2018.
\newblock ISBN 9780262039246.
\newblock URL \url{http://incompleteideas.net/book/RLbook2020.pdf}.

\bibitem[Burkard et~al.(1997)Burkard, Karisch, and Rendl]{Burkard1997}
Rainer~E Burkard, Stefan~E Karisch, and Franz Rendl.
\newblock Qaplib-a quadratic assignment problem library.
\newblock \emph{Journal of Global Optimization}, 10:\penalty0 391--403, 1997.

\bibitem[Whitley(2019)]{Whitley2019}
Darrell Whitley.
\newblock \emph{Next generation genetic algorithms: A user’s guide and tutorial}, volume 272.
\newblock Springer New York LLC, 2019.
\newblock \doi{10.1007/978-3-319-91086-4\_8}.

\bibitem[Azarbonyad and Babazadeh(2014)]{Azarbonyad2014}
Hosein Azarbonyad and Reza Babazadeh.
\newblock A genetic algorithm for solving quadratic assignment problem(qap).
\newblock \emph{arXiv}, 5 2014.
\newblock URL \url{http://arxiv.org/abs/1405.5050}.

\bibitem[Ferreira et~al.(2014)Ferreira, Quintas, Oliveira, Pereira, and Dias]{Ferreira2014}
João Ferreira, Artur Quintas, José~A. Oliveira, Guilherme A.~B. Pereira, and Luis Dias.
\newblock \emph{Solving the Team Orienteering Problem: Developing a Solution Tool Using a Genetic Algorithm Approach}, volume 223.
\newblock 2014.
\newblock ISBN 9783319009292.
\newblock \doi{10.1007/978-3-319-00930-8\_32}.
\newblock URL \url{http://link.springer.com/10.1007/978-3-319-00930-8\_32}.

\bibitem[Kramer(2017)]{Kramer2017}
Oliver Kramer.
\newblock \emph{Genetic Algorithm Essentials}, volume 679.
\newblock Springer International Publishing, 2017.
\newblock ISBN 978-3-319-52155-8.
\newblock \doi{10.1007/978-3-319-52156-5}.
\newblock URL \url{http://link.springer.com/10.1007/978-3-319-52156-5}.

\bibitem[Kleinberg et~al.(2008)Kleinberg, Slivkins, and Upfal]{Kleinberg2008}
Robert Kleinberg, Aleksandrs Slivkins, and Eli Upfal.
\newblock Multi-armed bandits in metric spaces.
\newblock In \emph{Proceedings of the fortieth annual ACM symposium on Theory of computing}, pages 681--690. ACM, 5 2008.
\newblock ISBN 9781605580470.
\newblock \doi{10.1145/1374376.1374475}.
\newblock URL \url{https://dl.acm.org/doi/10.1145/1374376.1374475}.

\bibitem[Kleinberg et~al.(2019)Kleinberg, Slivkins, and Upfal]{Kleinberg2019}
Robert Kleinberg, Aleksandrs Slivkins, and Eli Upfal.
\newblock Bandits and experts in metric spaces.
\newblock \emph{Journal of the ACM}, 66, 2019.
\newblock ISSN 1557735X.
\newblock \doi{10.1145/3299873}.

\bibitem[Whitley and Kauth(1988)]{Whitley1988}
Darrell Whitley and Joan Kauth.
\newblock Genitor: A different genetic algorithm.
\newblock In \emph{Proceedings of the 1988 Rocky Mountain Conference on Artificial Intelligence}, pages 189--214, 1988.

\bibitem[Thierens and Goldberg(1994)]{Thierens1994}
Dirk Thierens and David Goldberg.
\newblock Elitist recombination: An integrated selection recombination ga.
\newblock In \emph{IEEE Conference on Evolutionary Computation - Proceedings}, volume~1, pages 508--512. IEEE, 1994.
\newblock \doi{10.1109/icec.1994.349898}.

\bibitem[Matsui(1999)]{Matsui1999}
K.~Matsui.
\newblock New selection method to improve the population diversity in genetic algorithms.
\newblock In \emph{Proceedings of the IEEE International Conference on Systems, Man and Cybernetics}, volume~1, pages 625--630. IEEE, 1999.
\newblock \doi{10.1109/icsmc.1999.814164}.

\bibitem[Loshchilov et~al.(2011)Loshchilov, Schoenauer, and Sebag]{Loshchilov2011}
Ilya Loshchilov, Marc Schoenauer, and Michèle Sebag.
\newblock Not all parents are equal for mo-cma-es.
\newblock In \emph{International Conference on Evolutionary Multi-Criterion Optimization}, pages 31--45. Springer, Berlin, Heidelberg, 2011.

\bibitem[Belluz et~al.(2015)Belluz, Gaudesi, Squillero, and Tonda]{Belluz2015}
Jany Belluz, Marco Gaudesi, Giovanni Squillero, and Alberto Tonda.
\newblock Operator selection using improved dynamic multi-armed bandit.
\newblock In \emph{GECCO 2015 - Proceedings of the 2015 Genetic and Evolutionary Computation Conference}, pages 1311--1317. Association for Computing Machinery, Inc, 7 2015.
\newblock ISBN 9781450334723.
\newblock \doi{10.1145/2739480.2754712}.

\bibitem[DaCosta et~al.(2008)DaCosta, Fialho, Schoenauer, and Sebag]{DaCosta2008}
Luis DaCosta, Alvaro Fialho, Marc Schoenauer, and Michèle Sebag.
\newblock Adaptive operator selection with dynamic multi-armed bandits.
\newblock In \emph{Proceedings of the 10th annual conference on Genetic and evolutionary computation}, pages 913--920. ACM, 7 2008.
\newblock ISBN 9781605581309.
\newblock \doi{10.1145/1389095.1389272}.
\newblock URL \url{https://dl.acm.org/doi/10.1145/1389095.1389272}.

\bibitem[Gonçalves et~al.(2015)Gonçalves, Almeida, and Pozo]{Goncalves2015}
Richard~A. Gonçalves, Carolina~P. Almeida, and Aurora Pozo.
\newblock \emph{Upper Confidence Bound (UCB) Algorithms for Adaptive Operator Selection in MOEA/D}, volume 9018.
\newblock Springer International Publishing, 2015.
\newblock \doi{10.1007/978-3-319-15934-8\_28}.
\newblock URL \url{http://link.springer.com/10.1007/978-3-319-15934-8\_28}.

\bibitem[Li et~al.(2014)Li, Fialho, Kwong, and Zhang]{Li2014}
Ke~Li, Alvaro Fialho, Sam Kwong, and Qingfu Zhang.
\newblock Adaptive operator selection with bandits for a multiobjective evolutionary algorithm based on decomposition.
\newblock \emph{IEEE Transactions on Evolutionary Computation}, 18:\penalty0 114--130, 2 2014.
\newblock ISSN 1089778X.
\newblock \doi{10.1109/TEVC.2013.2239648}.

\bibitem[Vidal(2022)]{Vidal2022}
Thibaut Vidal.
\newblock Hybrid genetic search for the cvrp: Open-source implementation and swap* neighborhood.
\newblock \emph{Computers and Operations Research}, 140, 4 2022.
\newblock ISSN 03050548.
\newblock \doi{10.1016/j.cor.2021.105643}.

\bibitem[Jebari and Madiafi(2013)]{Jebari2013}
Khalid Jebari and Mohammed Madiafi.
\newblock Selection methods for genetic algorithms.
\newblock \emph{Int. J. Emerg. Sci}, 3:\penalty0 333--344, 2013.
\newblock ISSN 2222-4254.
\newblock URL \url{https://www.researchgate.net/publication/259461147}.

\bibitem[Shukla et~al.(2015)Shukla, Pandey, and Mehrotra]{Shukla2015}
Anupriya Shukla, Hari~Mohan Pandey, and Deepti Mehrotra.
\newblock Comparative review of selection techniques in genetic algorithm.
\newblock In \emph{2015 1st International Conference on Futuristic Trends in Computational Analysis and Knowledge Management, ABLAZE 2015}, pages 515--519. Institute of Electrical and Electronics Engineers Inc., 7 2015.
\newblock ISBN 9781479984336.
\newblock \doi{10.1109/ABLAZE.2015.7154916}.

\bibitem[Katoch et~al.(2021)Katoch, Chauhan, and Kumar]{Katoch2021}
Sourabh Katoch, Sumit~Singh Chauhan, and Vijay Kumar.
\newblock A review on genetic algorithm: past, present, and future.
\newblock \emph{Multimedia Tools and Applications}, 80:\penalty0 8091--8126, 2 2021.
\newblock ISSN 15737721.
\newblock \doi{10.1007/s11042-020-10139-6}.
\newblock URL \url{http://link.springer.com/10.1007/s11042-020-10139-6}.

\bibitem[Sanfeliu and Fu(1983)]{Sanfeliu1983}
Alberto Sanfeliu and King-Sun Fu.
\newblock A distance measure between attributed relational graphs for pattern recognition.
\newblock \emph{IEEE Transactions on Systems, Man, and Cybernetics}, SMC-13:\penalty0 353--362, 5 1983.
\newblock ISSN 0018-9472.
\newblock \doi{10.1109/TSMC.1983.6313167}.

\bibitem[Dang et~al.(2013)Dang, Guibadj, and Moukrim]{Dang2013}
Duc~Cuong Dang, Rym~Nesrine Guibadj, and Aziz Moukrim.
\newblock An effective pso-inspired algorithm for the team orienteering problem.
\newblock \emph{European Journal of Operational Research}, 229:\penalty0 332--344, 2013.
\newblock ISSN 03772217.
\newblock \doi{10.1016/j.ejor.2013.02.049}.
\newblock URL \url{http://dx.doi.org/10.1016/j.ejor.2013.02.049}.

\bibitem[Rey and Neuhauser(2011)]{Rey2011}
Denise Rey and Markus Neuhauser.
\newblock \emph{Wilcoxon-Signed-Rank Test}.
\newblock Springer Berlin Heidelberg, 2011.
\newblock \doi{10.1007/978-3-642-04898-2\_616}.

\bibitem[{SAS Institute Inc.}(2020)]{SAS2020}
{SAS Institute Inc.}
\newblock \emph{JMP(R) 16 Basic Analysis}.
\newblock SAS Institute Inc., 2020.

\bibitem[{Rochester Institute of Technology}(2019)]{RochesterInstituteofTechnology2019}
{Rochester Institute of Technology}.
\newblock {Research Computing Services}, 2019.
\newblock URL \url{{https://www.rit.edu/researchcomputing}}.

\end{thebibliography}

\end{document}